\documentclass[10pt, conference]{IEEEtran}
\IEEEoverridecommandlockouts
\usepackage{cite}
\usepackage{amsmath,amssymb,amsfonts}
\usepackage{textcomp}
\usepackage[table,xcdraw]{xcolor}
\def\BibTeX{{\rm B\kern-.05em{\sc i\kern-.025em b}\kern-.08em
    T\kern-.1667em\lower.7ex\hbox{E}\kern-.125emX}}

\usepackage{algorithm}

\usepackage[noend]{algpseudocode}
\usepackage{float}
\usepackage{caption}
\usepackage{multirow}  
\usepackage{amsmath}
\usepackage{titlesec}
\titlespacing*{\section}{0pt}{1.1\baselineskip}{\baselineskip}
\usepackage{cellspace, hhline}
\usepackage{mathtools}
\usepackage{subfig}

\usepackage{breqn}
\usepackage{hyperref}
\usepackage{amssymb}
\usepackage[export]{adjustbox}
\usepackage[symbol]{footmisc}
\def\algbackskip{\hskip-\ALG@thistlm}

\usepackage[switch]{lineno}
\newcommand\nnfootnote[1]{%
  \begin{NoHyper}
  \renewcommand\thefootnote{}\footnote{#1}%
  \addtocounter{footnote}{-1}%
  \end{NoHyper}
}

\newcommand\githubref[1]{%
  \renewcommand\thefootnote{}\footnote{#1}%
  \addtocounter{footnote}{-1}%
}

\makeatletter 
\newcommand{\algmargin}{\the\ALG@thistlm}
\makeatother
\algnewcommand{\parState}[1]{\State%
    \parbox[t]{\dimexpr\linewidth-\algmargin}{\strut\hangindent=\algorithmicindent \hangafter=1 #1\strut}}

\usepackage{xcolor,soul,lipsum}
\newcommand{\myul}[2][black]{\setulcolor{#1}\ul{#2}\setulcolor{black}}

\newcommand{\algorithmicbreak}{\textbf{break}}
\newcommand{\BREAK}{\State \algorithmicbreak}

\DeclareRobustCommand*{\IEEEauthorrefmark}[1]{%
\raisebox{0pt}[0pt][0pt]{\textsuperscript{\footnotesize #1}}%
}
\usepackage[nodisplayskipstretch]{setspace}
\begin{document}

\title{Towards Searching Efficient and Accurate Neural Network Architectures in Binary Classification Problems}

\author{\IEEEauthorblockN{Yigit Alparslan\textsuperscript{\textsection}\IEEEauthorrefmark{1}\IEEEauthorrefmark{*}, Ethan Jacob Moyer\textsuperscript{\textsection}\IEEEauthorrefmark{2},
Isamu Mclean Isozaki\textsuperscript{\textsection}\IEEEauthorrefmark{1}, Daniel Schwartz\IEEEauthorrefmark{1}\\ Adam Dunlop\IEEEauthorrefmark{3}, Shesh Dave\IEEEauthorrefmark{4}, Edward Kim\IEEEauthorrefmark{1}}

\IEEEauthorblockA{
\IEEEauthorrefmark{1}College of Computing \& Informatics, Drexel University, PA\\
\IEEEauthorrefmark{2}School of Biomedical Engineering, Drexel University, PA\\
\IEEEauthorrefmark{3}College of Arts \& Sciences, Drexel University, PA\\
\IEEEauthorrefmark{4}College of Engineering, Drexel University, PA\\
Email: \{ ya332, ejm374, imi25, des338, ajd393, sd3536, ek826 \}@drexel.edu}}

\maketitle
\begingroup\renewcommand\thefootnote{\textsection}
\footnotetext{ These co-first authors contributed equally.}
\endgroup

\begin{abstract}

In recent years, deep neural networks have had great success in machine learning and pattern recognition. Architecture size for a neural network contributes significantly to the success of any neural network. In this study, we optimize the selection process by investigating different search algorithms to find a neural network architecture size that yields the highest accuracy. We apply binary search on a very well-defined binary classification network search space and compare the results to those of linear search. We also propose how to relax some of the assumptions regarding the dataset so that our solution can be generalized to any binary classification problem. We report a 100-fold running time improvement over the naive linear search when we apply the binary search method to our datasets in order to find the best architecture candidate. By finding the optimal architecture size for any binary classification problem quickly, we hope that our research contributes to discovering intelligent algorithms for optimizing architecture size selection in machine learning.


\end{abstract}

\nnfootnote{\IEEEauthorrefmark{*} Corresponding author}
\githubref{All source code is open-sourced at \href{https://github.com/drexelai/binary-search-in-neural-nets}{\color{blue} \myul[blue] {GitHub.}}}

\section{Introduction}
\IEEEPARstart{I}{n} recent decades, deep neural networks (DNNs) have seen many breakthroughs to achieve or even exceed human-level performance on difficult classification and recognition tasks. 
Breakthroughs in many challenging applications, such as speech recognition \cite{hinton2012} \cite{alparslanspeech2020}, image recognition \cite{lecun1998} 
\cite{krizhevsky2012} \cite{greenstadt2020}, genomic classification \cite{moyer2020machine}, machine translation \cite{sutskever2014} \cite{bahdanau2015}, have been achieved due to intelligent architectures that have been designed for the task at hand.

As described in \cite{le2016}, one such architectural breakthrough was in computer vision to predict objects in images by AlexNet~\cite{krizhevsky2012}, VGGNet \cite{simonyan2014}, GoogleNet \cite{szedegy2015}, and
ResNet\cite{zhang2016} which replaced the previously used architecture designs that were based on features such as SIFT \cite{lowe1999} and HOG \cite{dalal2005}. Designing architectures for a specific problem and dataset enabled greater success in many other fields as well such as voice recognition~\cite{speechrecognitionbreakthrough}. However, models started to require many small design choices, increasingly sophisticated details, and many hyperparameters - parameters chosen by a user. These hyperparameters, such as the number of hidden layers, the number of nodes at each layer etc. affects the accuracy, model training duration, and architecture size directly. Therefore, recent years have seen a surge in interest in the Neural Architecture Search (NAS) field. NAS generally dictates generating a search space with all the artificial neural networks that can be designed and optimized, and then a search strategy is implemented to go over and find the best candidate among all the neural network architectures in that space. A search strategy can be optimized to skip similar candidates, and find the most accurate architectures. Zoph et al. \cite{rloutperformcifar}  outperformed the best manually designed architecture for the CIFAR-10 dataset by finding a model architecture 1.05x faster and with 0.09\% better accuracy.  

In NAS, even though the accuracy of the model is the most important metric, other metrics such as memory consumption, training time, inference time, model size could also be important when choosing a search strategy.  

In this study, we search for architecture sizes that would give the highest accuracy and lowest training time for a given, well-defined architecture size search space with certain assumptions. Our aim is to look at the architecture size as a hyperparameters and propose a framework for discovering the most optimal architecture size for a given problem. We specifically consider the number of hidden layers and the number of neurons at each layer for a given problem and search the most optimal settings in a deep neural network (DNN). We consider a limited set of networks that satisfy the binary classification problem.  Thus, the output for all the networks only have a single output neuron.  We use binary search and linear search as a way of finding the optimal architecture sizes for problems of this nature.  In our experiments, we investigate the Titanic dataset and a Customer Churn dataset to compare and apply our findings.  We treat the linear search method as a baseline and report qualitative and quantitative improvements via binary search over the baseline. 

This paper is organized such that \autoref{relatedwork} discusses related work, \autoref{methodology} discusses the implemented search algorithms, \autoref{experimentandresults} reports the experiments and results, \autoref{conclusion} concludes the paper by going over the important findings one more time, and finally \autoref{application} discusses future work.

\section{Related Work} \label{relatedwork}

Architecture size has long been considered a hyperparameter that a user picks randomly or heuristically. Yet, this hyperparameter impacts the model accuracy for a given problem significantly. Recent years have shown several studies in which hyperparameters are optimized \cite{bergstra2011}\cite{bergstrabengio2012} \cite{snoek2012} \cite{saxena2016}. Such studies have been limited to fixed-size models when searching for the optimal hyperparameters. 

Zoph and Le \cite{le2016} relaxed the fixed-space assumption using reinforcement learning. Shen et al. \cite{shenetal2019} has focused on finding the optimal architecture size for binary neural networks, which are neural networks where the weights consist of only +1 and -1 values. These two studies have proposed new frameworks for determining the architecture size of neural networks in a systematic way rather than leaving such choice seemingly ambiguous. However, such studies have taken into account different assumptions regarding the models they design when applying each approach. to limit the search space. For example, Shen et al\cite{shenetal2019} added the constraint of architecting binary neural network to only look most compact at the neural networks.  
This constraint that they added is only slightly similar to our assumptions in this work. Instead of a forced binarization on the weights, we simply assume that our classification problems are ones that can be implemented with binary outputs. In other words, we only investigate our search algorithms to datasets that can be solved with a model where the last layer has one node. Additionally, Alparslan et al. \cite{sparsitypaper} worked on using sparsity as a heuristics to find the architecture candidates that would give the most sparse as well as accurate models.
 
\section{Methodology} \label{methodology}

Because of the fact that DNNs require training and testing on typically large datasets, it becomes increasingly difficult and time consuming to determine optimal hyperparameters. 

In this work, we define model evaluation as the end-to-end training and testing of a model architecture with a constant training and testing dataset, which is specially outlined in \autoref{datasetassumption}. 

\subsection{Assumptions}

Often in machine learning, algorithms are implemented with general assumptions in order to simplify the problem. For instance, the Naive Bayes classification algorithm relies on the assumption that all features are perfectly independent of each other given a certain class \cite{rish2001empirical}. Intuitively, this is a poor assumption because features of a given class interact in some way or another. Regardless, Naive Bayes has proved time and time again to be an excellent classifier in text classification \cite{ting2011naive} \cite{kim2006some}. Additional example is the fact that most neural networks  use gradient descent algorithm as optimization algorithm. The condition to apply gradient ascent is that we have to assume the function is continuously convex and differentiable\cite{chongzakgradientdescent}. However, cost functions that are optimized by neural networks might not meet this condition. So, in theory, those neural networks don't guarantee globally optimal solutions, but in practice, neural networks converge to a local minimum point as proven by Zhong et al. \cite{zhongetal2017a} \cite{zhongetal2017b} and Zhang et al. \cite{zhangetal2018}. Just like the cases of Naive Bayes and Gradient Descent, we also make assumptions in our paper that help explain our method, but do not have to hold in order to achieve good results in practice. 

This binary search method that we outline in this paper can be considered as follows: a trade-off between speed and accuracy. If the assumptions are met, it finds the globally optimal solution quickly (see \autoref{table:cuspsearching}). If the assumptions are not met, it finds, at the very least, a local optimum (see \autoref{table:modelsearching}). In our framework, we describe \textbf{the following three assumptions} that limit the problem space upon which we apply our binary search method. 

\subsubsection{Network Architecture Assumption}
We will be modeling our classification problem with an input layer, one hidden layer, and one output layer as shown below.
\begin{figure}[H]
    \centering
    \includegraphics[width=0.6\linewidth]{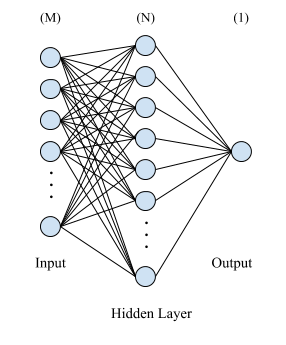}
    \caption{A generalized one-layer deep neural network with M input neurons, N hidden layer neurons, and one output neuron.}
    \label{fig:my_label}
\end{figure}

\subsubsection{Accuracy Distribution Assumption}
Our second general assumption is that the accuracy distribution is uni-modal with respect to the number of units, $N$, in the hidden layer. There exists only one global maximum when accuracies are plotted against each architecture with a hidden layer of dimension between $1$ and $n$ and accuracy values increase from both sides until global maximum. 

\subsubsection{Dataset Assumptions} \label{datasetassumption}
With respect to our dataset, we assume that there will be $M$ number of inputs but only $1$ output unit in the output layer. As a result, we are simplifying this to only binary classification problems. We suspect that specifically our linear search method in \autoref{linearsearch} will be more efficient for smaller input spaces, whereas our binary search method in \autoref{binarysearch} will be more efficient for larger input spaces. It follows that we can explore the relative speeds of these two methods under our respective assumptions in order to determine a general threshold for when one method should be used over the other.
\subsection{Datasets} \label{datasets}
\subsubsection{Churn dataset}
Churn dataset is made public by Drexel Society of Artificial Intelligence \cite{churndataset}. It has 14 columns and 10000 rows where each row has information for a business that uses a cloud service and each column represents one feature regarding the customer. The dataset has specifically 14 columns but 3 of them are row number, customer id and company name which are excluded during the training process because they do not have meaning for the model. The remaining columns are features to indicate the revenue of the customer, contract duration of the customer, whether the customer has raised a ticket or renewed the contract before etc. The model that is trained on this dataset predicts whether the customer will leave the service contract or not. So, the label column is either 0 or 1 to indicate if the customer will renew the service. We train a model that has 1 input layer that consist of 11 nodes, one hidden layer that consists of D nodes and 1 output layer that consist of 1 node. Next, we find the node count, D, that would give the maximum accuracy in our model- first, via linear search and, second, via (modified) binary search. We also investigate whether the solution provided by the binary search is a globally optimal solution. In other words, if there is a global maximum in all the accuracies given by all the architecture candidates, the binary search, just like the linear search should find the globally maximum accuracy value.

\subsubsection{Titanic dataset}
Titanic dataset is a dataset made public by Kaggle \cite{titanickaggledataset}. It has 14 columns and 1310 rows where each row has information for a passenger on Titanic and each column is one of many features. We use 11 features such as fare amount, gender, ticket class, cabin type etc from the dataset and the label for each sample that the model predicts is either 0 or 1 to indicate if the person survived or not. We train a model that has 1 input layer that consist of 11 nodes, one hidden that consists of D nodes and 1 output layer that consist of 1 node. The methods we follow is to find the value of D that would give the maximum accuracy in our model first via linear search and second via (modified) binary search. 

\subsection{Search}
For each model, we have a varying hidden layer of dimension, D, between the input layer and the output layer. We sweep D from 1 to $n$ to find the maximum accuracy which we treat as our baseline. In our studies, $n$=1000 proved to be effective because we have not realized any benefits of going more than three orders of magnitude larger than the input dimensions (11 nodes) in our problems.  Such linear search outlined in \autoref{linearsearch} takes exactly $n$ iterations meaning we need to train and test $n$ models with different architectures to find the highest accuracy. We then assume the accuracies are fit to a distribution with a single global maximum and apply binary search to skip some number of model architectures which then helps us the training time reduce by 100 fold and achieves a maximum accuracy in around 10 iterations. We discuss ways of applying binary search to this problem in section \ref{binarysearch}.

\subsubsection{Linear Search} \label{linearsearch}
The usage of linear search is considered baseline in our study for the generalized dataset that follows the accuracy distribution as mentioned in \autoref{datasetassumption} and the two datasets described in \autoref{datasets}. Linear search is a brute force method because one must iterate over all the elements and checking if the current element is greater than the current maximum element that has been observed so far in the search. We pick a range to constrain the search space so that finding the number of hidden layer units is a bounded problem instead of an unbounded problem. The time complexity for finding the maximum using linear search is O(n). It should be noted that although the search is greedy, this method is able to find the optimal hyperparameters perfectly in each search. Algorithm~\ref{alg:linearserachalg} formalizes this linear search algorithm. 

\begin{algorithm}[H] 
\caption{Finding maximum accurate architecture size in a neural network via linear search}
\label{alg:linearserachalg}
\begin{algorithmic}[1]

\Procedure{maximizeAccuracyLinearly}{}
\State $\textit{n} \gets \text{upper bound of dimension in a hidden layer}$
\State $\textit{max\_accuracy} \gets 0$
\State $\textit{max\_network\_size} \gets 1$
\For{$current\_size$ := 1 to $n$} 
\State $\textit{current\_accuracy} \gets \newline \textit{Pipeline(model(current\_size))}$
\If {$\textit{current\_accuracy} > \textit{max\_accuracy}$}
\State $max\_accuracy \gets current\_accuracy$
\State $max\_network\_size \gets current\_size$
\EndIf
\EndFor
\Return $(max\_accuracy, max\_network\_size)$
\EndProcedure
\end{algorithmic}
\end{algorithm}

\subsubsection{Binary Search} \label{binarysearch}

Given our assumptions, the premise of the binary search method is to implement a way to determine from which side we are approaching the cusp. Because of the fact that the slope of the distribution is generally monotonically increasing in magnitude in approaching the maximum accuracy, we model the index based on the sign of the recorded slope and the previously recorded slopes in the search.

\begin{figure}[!t]
     \centering
     \subfloat[Binary hyperparameter search. Initial conditions.]
     {\label{fig:binarysearch0}{\includegraphics[width=1\linewidth]{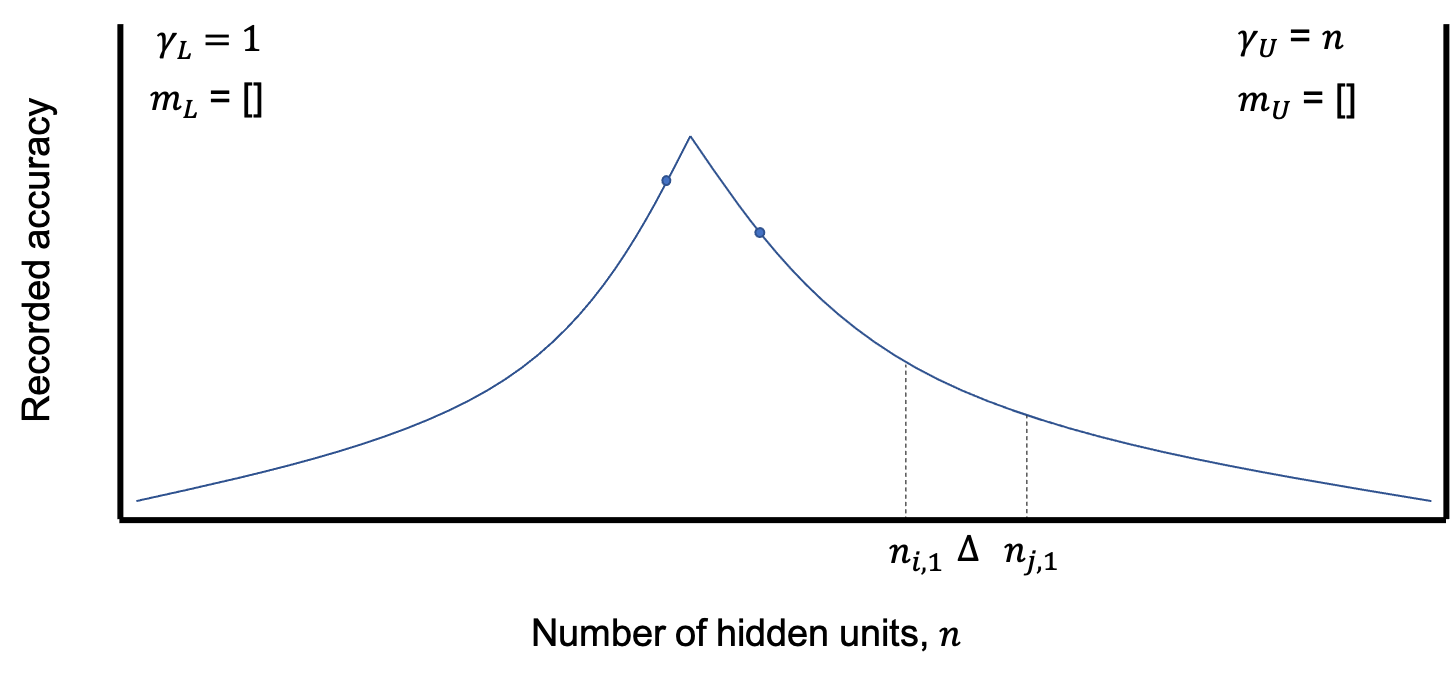}}}\hfill
    \subfloat[Binary hyperparameter search. First comparison.]{\label{fig:binarysearch1}{\includegraphics[width=1\linewidth]{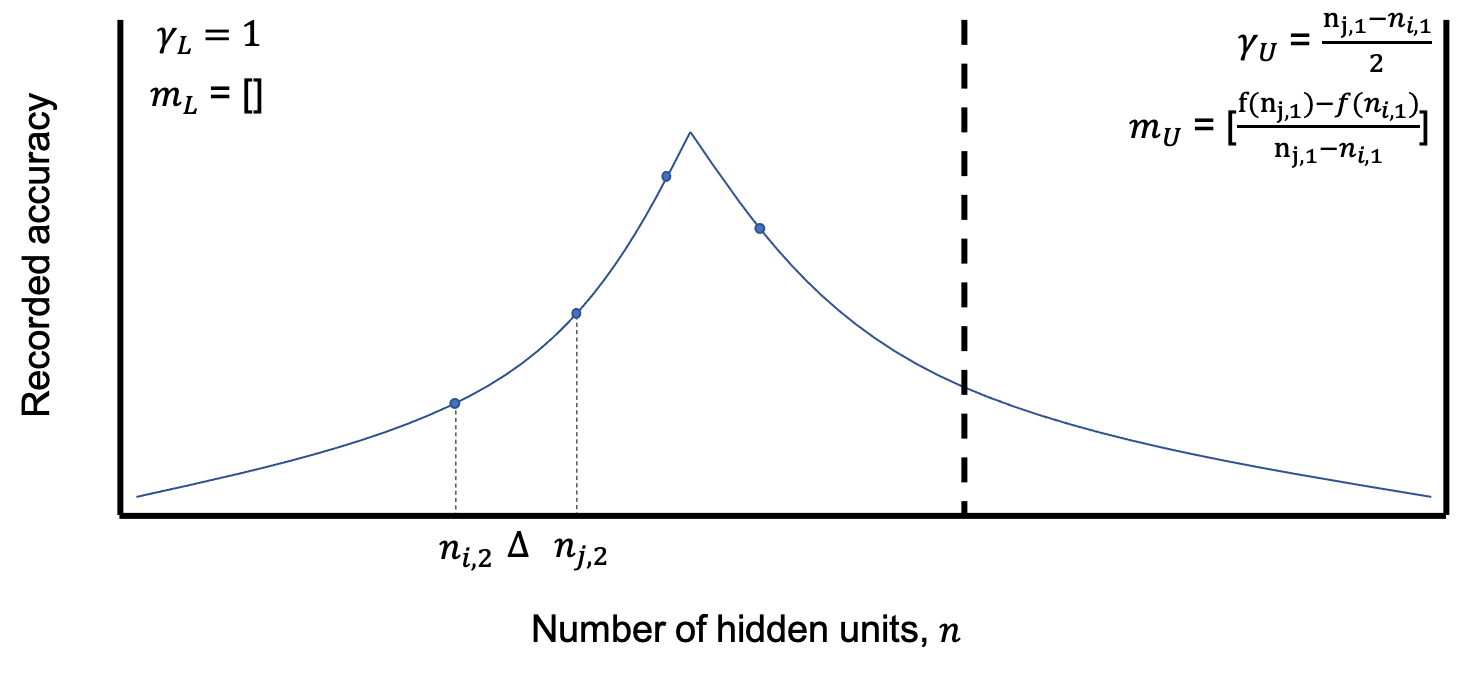}}}
    \caption{Binary hyperparameter search. A change in the sign of the slope indicates that there is a global maximum. Binary search rejects the side where there is no global maximum until it finds the global maximum. }
    \label{fig:binarysearch}
\end{figure}

In general, binary search can halve the input space at every comparison, because the search takes advantage of the fact that the dataset is sorted. During binary search, performing one comparison to check whether the current candidate is the target in a sorted dataset can eliminate the current candidate during that comparison as well as all the other candidates worse than the current one because the dataset is sorted. In other words, linear search is removing one candidate at each step, whereas binary search is removing half of the current input space at each step. We adapt this binary search idea and perform it in a similar way using slopes. Therefore, each comparison will take at least two model evaluations for each comparison. We introduce a variable $\delta$ that models the distance between each model evaluation taken at $n_i$ and $n_j$, which represent the current number of units on which we base our comparison and search.  \autoref{fig:binarysearch} models the relationship between these three variables in the search algorithm.

We then define  $\gamma_L$ and $\gamma_U$, which represent the minimum and maximum number of units in our search space, respectively. These variables will be used to keep track of the lower and upper bounds of our search. As shown in \autoref{fig:binarysearch0}, these two variables are set to $1$ and $n$ by default. If the search continues appropriately, these will approach the cusp from either side.

Moreover, we define lists $m_L$ and $m_U$ for the previously recorded slopes for the lower and upper bound side of the maximum cusp, respectively. Initially, these lists are empty as displayed in Figure \autoref{fig:binarysearch0}. As the search advances, the previously recorded slopes will be appended to either list depending on from which side the slope was taken.

After initializing these variables, the search begins by performing two model evaluations at $n_i$ and $n_j$ in Figure \autoref{fig:binarysearch0}. This would result in a negative slope, which indicates that the upper bound conditions are changed as shown in Figure \autoref{fig:binarysearch1}. The upper bound $\gamma_U$ is set to the $n$ from which the slope was estimated, and the slope is appended to the list containing previously recorded slopes on the right side of the cusp, $m_U$. As displayed in this figure, the search continues to the opposite side of the search space. 

After each slope is calculated, we want to know whether there is enough evidence to suggest that there exists a maximum either at or near that recorded slope. We determine the probability of a maximum by evaluating a posterior shown in \autoref{posteriorequ}.\\\\

\textbf{Posterior:}\\
\begin{equation}\label{posteriorequ}
\centering
\begin{aligned}
    P(maximum \mid m, \gamma_{L}, \gamma_{U}, \delta) = P(y{=}m {\mid} maximum) \times \\
    P(maximum {\mid} \gamma_{L}, \gamma_{U}, \delta)
\end{aligned}
\end{equation}

\begin{figure}[!ht]
    \centering
    \includegraphics[width=0.9\linewidth]{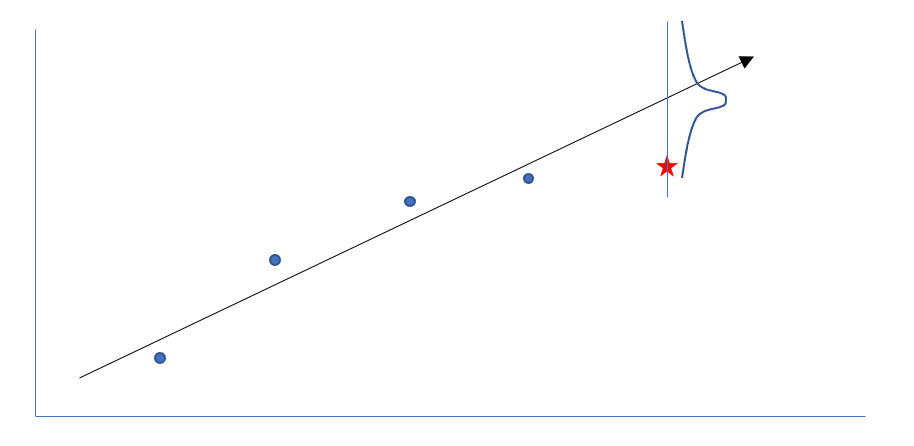}
    \caption{Binary search method. When a normal distribution is fit on the next point, points that are far from the mean by 2 standard deviations will be assigned a very small prior probabilities.}
    \label{likelihoodmodel}
\end{figure}

This probability can be broken down into a likelihood in \autoref{likelihoodequ} and a prior in \autoref{priorequ}. The likelihood represents the probability of observing a maximum given a history of maximums. This is implemented by modeling the next maximum with a linear regression with respect to the previously recorded maximums. This next expected maximum is then modeled as a normal distribution with a standard deviation equal to that of the regression as shown in \autoref{likelihoodmodel}.

\textbf{Likelihood:}
\begin{equation}\label{likelihoodequ}
        P(y=m \mid maximum) = N(\hat{y} = \beta_{0} + \beta_{1} * x, \sigma) 
\end{equation}

\textbf{Prior:}
\begin{equation}\label{priorequ}
\centering
\begin{aligned}
    P(maximum \mid \gamma_{L}, \gamma_{U}, \delta) = \frac{\delta}{\gamma_{U} - \gamma_{L}}
\end{aligned}
\end{equation}

The prior calculated in \autoref{priorequ} is simply the probability of discovering the maximum at random between two points based on the initial $\gamma_{L}$ and $\gamma_{U}$. 
Additionally, calculating prior probability takes O(1) time.

Algorithm~\ref{alg:binarsearchalg} formalizes the binary search algorithm. 

\begin{algorithm}[H] 
\caption{Finds maximum accurate architecture size in a neural network via binary search}
\label{alg:binarsearchalg}
\begin{algorithmic}[1]
\Procedure{maximizeAccuracyBinary}{}
\State $\textit{n} \gets \text{upper bound of dimension in a hidden layer}$
\State $\textit{$\gamma_L$} \gets \text{lower bound of dimension in a hidden layer}$
\State $\textit{$\gamma_U$} \gets \textit{n}$
\State $\textit{$\delta$} \gets \textit{sufficient distance between $n_i$ and $n_j$}$
\State $\textit{$\alpha$} \gets \textit{threshold probability that a maximum exists}$
\State $\textit{$m_L$} \gets []$
\State $\textit{$m_U$} \gets []$
\State $\textit{$mid_L$} \gets []$
\State $\textit{$mid_U$} \gets []$
\State $\textit{max\_accuracy} \gets 0$
\State $\textit{max\_network\_size} \gets 1$
\While{$\gamma_L$ $<=$ $\gamma_U$} 
\State $\textit{current\_size} \gets \frac{\gamma_{U} - \gamma_{L}}{2}$

\State $\textit{current\_slope} \gets \textit{getSlope(current\_size, $\delta$)}$

\If {$\textit{current\_slope} > \textit{0}$}
\State $\textit{$m_L$} \gets \textit{append($m_L$, current\_slope)}$
\State $\textit{$mid_L$} \gets \textit{append($mid_L$, current\_size)}$
\If {$\textit{getPosteriorProbability($n$, $\delta$, $m_L$,$mid_L$} \newline \indent \indent \textit{side=1)}$ $>$ $\alpha$}
\State $\textit{max\_network\_size} \gets \textit{$mid_L$[len($mid_L$)]}$
\BREAK
\Else
\State $\textit{$\gamma_L$} \gets \textit{current\_size}$
\EndIf
\Else
\State $\textit{$m_U$} \gets \textit{append($m_U$, current\_slope)}$
\State $\textit{$mid_U$} \gets \textit{append($mid_U$, current\_size)}$
\If {$\textit{getPosteriorProbability($n$, $\delta$, $m_U$, $mid_U$,} \newline
\indent \indent \textit{side=2)}$ $>$ $\alpha$}
\State $\textit{max\_network\_size} \gets \textit{$mid_U$[len($mid_L$)]}$
\BREAK
\Else
\State $\textit{$\gamma_U$} \gets \textit{current\_size}$
\EndIf
\EndIf
\EndWhile
\State $\textit{max\_accuracy} \gets \textit{Pipeline(model(max\_network\_size))}$
\Return $(max\_accuracy, max\_network\_size)$
\EndProcedure
\end{algorithmic}
\end{algorithm}

In algorithm~\ref{alg:binarsearchalg}, line 32, the $Pipeline$ that we have is a set of scripts to take a hidden layer size as input, generate a $model$ and run training and testing on it. Line 1 and 2 in algorithm~\ref{alg:getslope} also uses the same $Pipeline$ to automate the training and testing for architecture candidates that iterate over. Additionally, calculating the the posterior likelihood for being a maximum for each candidate in the search space means that we need to set up a threshold to accept the current candidate as the best candidate and stop the algorithm. From empirical evidence in two datasets that we studied, 2 standard deviations ($\sigma$) distance from the mean for a normal distribution is enough to accept any candidate as the architecture candidate with the best accuracy. 

\begin{algorithm}[H]
\caption{Evaluates posterior probability of whether a maximum exists between two points}
\begin{algorithmic}[1]
\Procedure{getPosteriorProbability}{$n$, $\delta$, $m$, $mid$, $side$}

\If {$m$ == 1}
\State $\textit{likelihood} \gets 0.5$
\ElsIf {$m$ == 2}
\State $\textit{likelihood} \gets norm(m[0], sigma).cdf(m[1])$
\Else

\State $\textit{x} \gets mid[:-1]$
\State $\textit{y} \gets m[:-1]$

\State $\textit{model} \gets LinearRegression(x, y)$

\State $\textit{y\_pred} \gets model.predict(x)$
\State $\textit{sigma} \gets std(y\_pred)$

\State $\textit{$y_i$} \gets model.predict(mid[-1])$

\State $\textit{likelihood} \gets norm(y_i, sigma).cdf(mid[-1])$
\EndIf

\If{side == 1}
\State $\textit{likelihood} \gets 1 - likelihood$
\EndIf

\State $\textit{prior} \gets \frac{\delta}{n - 1}$

\State $\textit{posterior} \gets \textit{likelihood} * \textit{prior}$
\Return $posterior$
\EndProcedure
\end{algorithmic}
\end{algorithm}

This can be attributed to the fact that 95\% of all data in a normal distribution can be fit into 2 standard deviation within the mean. Such usage of a predetermined acceptance threshold would mean to stop the algorithm early and save CPU time. A second choice would be to run the algorithm to completion with no early stopping until each candidate checked by the algorithm gets assigned a posterior likelihood and then pick the one with the highest posterior likelihood.

\begin{algorithm}[H]
\caption{Returns slope of secant line between two given architecture sizes separated by $\delta$}
\label{alg:getslope}
\begin{algorithmic}[1]

\Procedure{\textit{getSlope}}{$current\_size, \delta$}

\State $accuracy_i \gets \textit{Pipeline(model(current\_size +$\frac{\delta}{2}$))}$ 

\State $accuracy_j \gets \textit{Pipeline(model(current\_size -$\frac{\delta}{2}$))}$

\Return $\frac{accuracy_i - accuracy_j}{\delta}$
\EndProcedure
\end{algorithmic}
\end{algorithm}

\begin{table*}[t]
\centering
\setlength{\belowcaptionskip}{15pt}
\caption{Experiment results displaying the difference between linear search and binary search on the sample search architecture search space.}
\label{table:cuspsearching}
\begin{tabular}{|c|c|c|c|c|}
\hline
\textbf{Search} & \multicolumn{2}{c|}{\textbf{Number of evaluations}} & \multicolumn{2}{c|}{\textbf{Evaluation Error}} \\
\cline{2-5}
\textbf{Methods} & \textbf{\textit{Average}}        & \textbf{\textit{Standard Deviation}}        & \textbf{\textit{Average}}      & \textbf{\textit{Standard Deviation}}     \\ \hline
\textbf{Linear}                        & 1000           & 0                         & 0            & 0                      \\ \hline
\textbf{Binary}                        & 6.458          & 0.9477                    & 1.152        & 0.8915                 \\ \hline
\end{tabular}
\end{table*}

\begin{table*}[t]
\centering
\setlength{\belowcaptionskip}{5pt}
\caption{Linear and Binary Search were used to find the best models for both datasets and their training and testing accuracies are reported as well as the their sizes and the time spent on searching them. Every candidate model was trained over 15 epochs when searching. Finding the best model via linear search took 1000 steps for both models, or 1000$\times$15 epochs each. Finding the best model via binary search took 11 steps for the titanic model and 10 steps for the churn model.}
\label{table:modelsearching}
\begin{tabular}{|c|c|c|c|c|c|}
\hline
\multicolumn{1}{|c|}{\textbf{\parbox[t]{5cm}{Model Type, Iteration, Dropout}}} &
  \textbf{\parbox[t]{2cm}{Linear Search\\Duration}} & \textbf{\parbox[t]{2cm}{Binary Search\\Duration}} & \textbf{\parbox[t]{2cm}{Best Model's\\Training\\Accuracy}} & \textbf{\parbox[t]{2cm}{Best Model's\\ Testing\\ Accuracy}} & \textbf{\parbox[t]{2cm}{Best Model's\\Architecture}} \\ \hline

\multicolumn{1}{|c|}{\textbf{Titanic Model\_100 (w/o Dropout) (\ref{fig:titanic_accuracy_100})}} & 100 steps  & 7 steps & 81.9\% & 79.6\%  & 24 nodes \\ \hline

\multicolumn{1}{|c|}{\textbf{Titanic Model\_100 (w/ Dropout) (\ref{fig:titanic_accuracy_100_with_dropout})}} & 100 steps  & 7 steps & 78.1\% & 80.7\%  & 61 nodes \\ \hline

\multicolumn{1}{|c|}{\textbf{Titanic Model\_1000 (w/o Dropout) (\ref{fig:titanic_accuracy_1000})}} & 1000 steps  & 11 steps & 81.9\% & 79.6\%  & 787 nodes \\ \hline

\multicolumn{1}{|c|}{\textbf{Titanic Model\_1000 (w/ Dropout) (\ref{fig:titanic_accuracy_1000_with_dropout})}} & 1000 steps  & 9 steps & 83.4\% & 78.5\%  & 410 nodes \\ \hline

\multicolumn{1}{|c|}{\textbf{Churn Model\_100 (w/o Dropout) (\ref{fig:churn_accuracy_100})}} & 100 steps  & 6 steps & 94.7\% & 77.5\%  & 1 node \\ \hline

\multicolumn{1}{|c|}{\textbf{Churn Model\_100 (w/ Dropout) (\ref{fig:churn_accuracy_100_with_dropout})}} & 100 steps & 7 steps & 96.2\% & 78.4\%  & 64 nodes \\ \hline

\multicolumn{1}{|c|}{\textbf{Churn Model\_1000 (w/o Dropout) (\ref{fig:churn_accuracy_1000})}} & 1000 steps & 10 steps & 86.8\% & 80.2 \% & 804 nodes \\ \hline

\multicolumn{1}{|c|}{\textbf{Churn Model\_1000 (w/ Dropout) (\ref{fig:churn_accuracy_1000_with_dropout})}} & 1000 steps  & 9 steps & 83.1\% & 78.2\%\%  & 511 nodes \\ \hline

\end{tabular}
\end{table*}

\section{Experiment Results and Observations}
\label{experimentandresults}

\begin{figure}[bp!]
    \centering
    \includegraphics[width=0.8\linewidth]{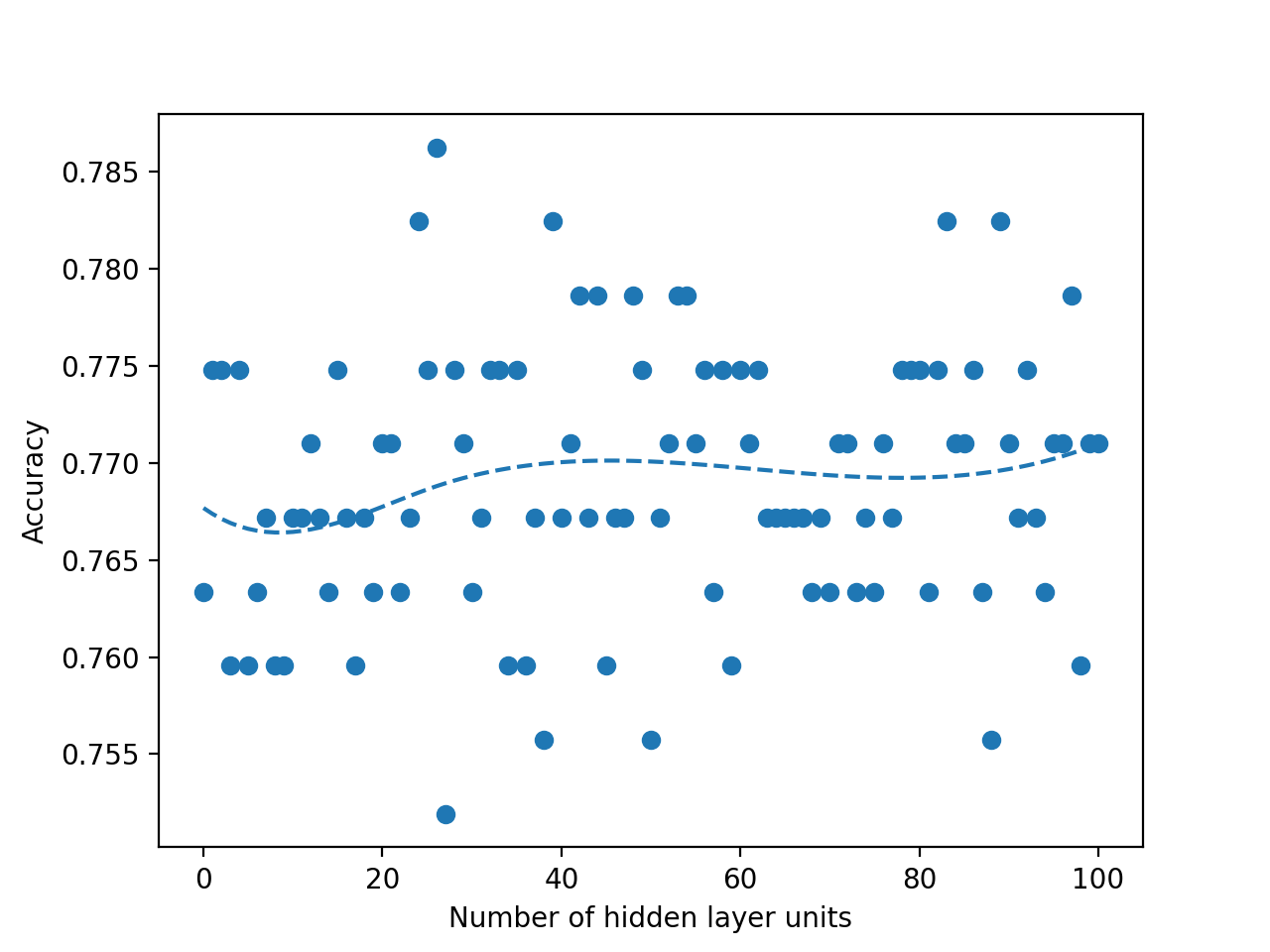}
    \setlength{\belowcaptionskip}{-15pt}
    \caption{Accuracy vs Number of Hidden Layer Units in the \textbf{Titanic Model}. \textbf{100} different models were created and trained with same input and output layer and different hidden layer. The resulting curve of model accuracies is where the linear and binary search are applied.}
    \label{fig:titanic_accuracy_100}
\end{figure}

\begin{figure}[bp!]
    \centering
    \includegraphics[width=0.8\linewidth]{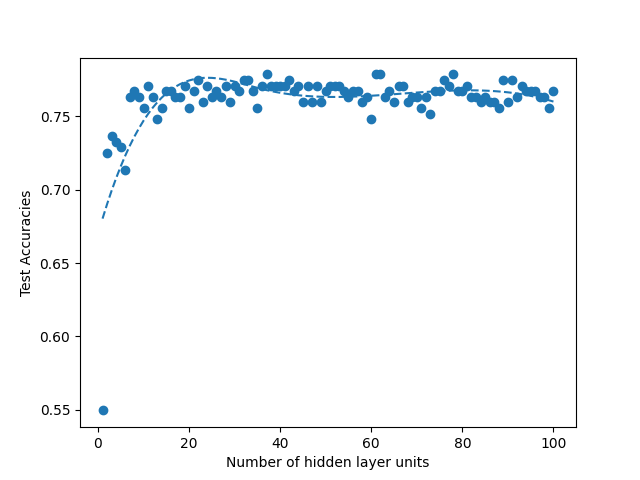}
    \setlength{\belowcaptionskip}{-15pt}
    \caption{Accuracy vs Number of Hidden Layer Units in the \textbf{Titanic Model}. \textbf{100} different models with \textbf{Dropout layer} applied were created and trained with same input and output layer and different hidden layer. The resulting curve of model accuracies is where the linear and binary search are applied.}
    \label{fig:titanic_accuracy_100_with_dropout}
\end{figure}

\begin{figure}[bp!]
    \centering
    \includegraphics[width=0.8\linewidth]{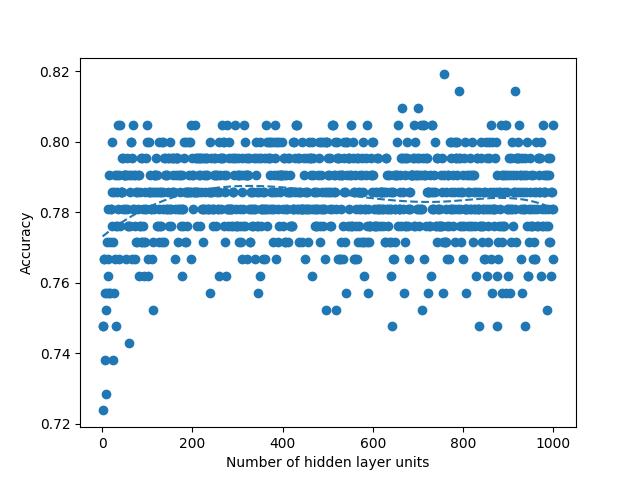}
    \caption{Accuracy vs Number of Hidden Layer Units in the \textbf{Titanic Model}. \textbf{1000} different models were created and trained with same input and output layer and different hidden layer. The resulting curve of model accuracies is where the linear and binary search are applied.}
    \label{fig:titanic_accuracy_1000}
\end{figure}

\begin{figure}[bp!]
    \centering
    \includegraphics[width=0.8\linewidth]{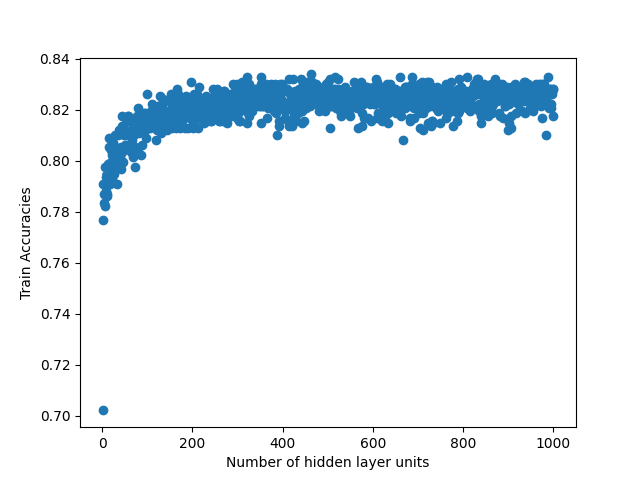}
    \setlength{\belowcaptionskip}{-15pt}
    \caption{Accuracy vs Number of Hidden Layer Units in the \textbf{Titanic Model}. \textbf{1000} different models with \textbf{Dropout layer} applied were created and trained with same input and output layer and different hidden layer. The resulting curve of model accuracies is where the linear and binary search are applied.}
    \label{fig:titanic_accuracy_1000_with_dropout}
\end{figure}

\begin{figure}[hp!]
    \centering
    \includegraphics[width=0.8\linewidth]{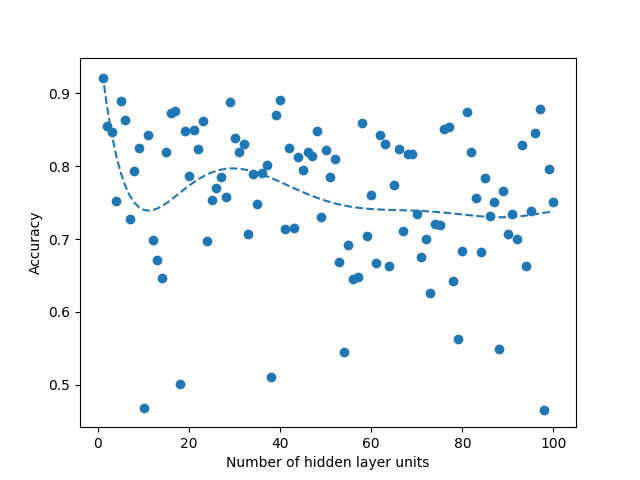}
    \setlength{\belowcaptionskip}{-15pt}
    \caption{Accuracy vs Number of Hidden Layer Units in the \textbf{Churn Model}. \textbf{100} different models were created and trained with same input and output layer and different hidden layer. The resulting curve of model accuracies is where the linear and binary search are applied. }
    \label{fig:churn_accuracy_100}
\end{figure}

\begin{figure}[hp!]
    \centering
    \includegraphics[width=0.8\linewidth]{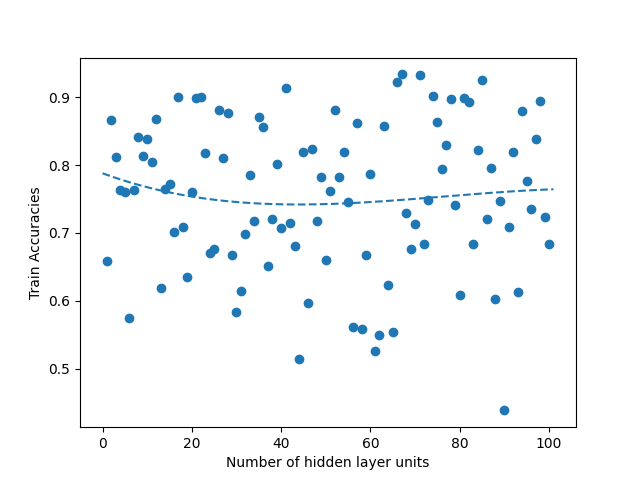}
    \setlength{\belowcaptionskip}{-15pt}
    \caption{Accuracy vs Number of Hidden Layer Units in the \textbf{Churn Model}. \textbf{100} different models with \textbf{Dropout layer} applied were created and trained with same input and output layer and different hidden layer. The resulting curve of model accuracies is where the linear and binary search are applied.}
    \label{fig:churn_accuracy_100_with_dropout}
\end{figure}

\begin{figure}[hp!]
    \centering
    \includegraphics[width=0.8\linewidth]{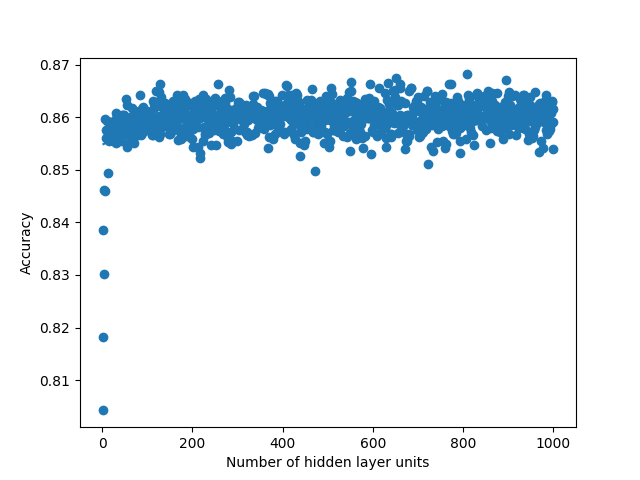}
    \caption{Accuracy vs Number of Hidden Layer Units in the \textbf{Churn Model}. \textbf{1000} different models were created and trained with same input and output layer and different hidden layer. The resulting curve of model accuracies is where the linear and binary search are applied. }
    \label{fig:churn_accuracy_1000}
\end{figure}

\begin{figure}[tp]
    \centering
    \includegraphics[width=0.8\linewidth]{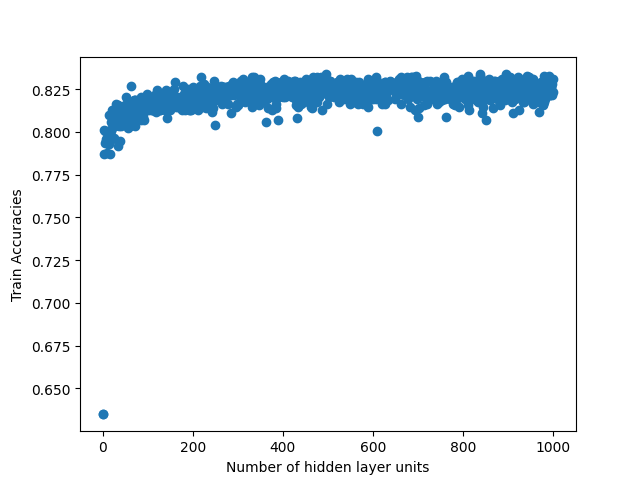}
    \setlength{\belowcaptionskip}{-15pt}
    \caption{Accuracy vs Number of Hidden Layer Units in the \textbf{Churn Model}. \textbf{1000} different models with \textbf{Dropout layer} applied were created and trained with same input and output layer and different hidden layer. The resulting curve of model accuracies is where the linear and binary search are applied.}
    \label{fig:churn_accuracy_1000_with_dropout}
\end{figure}

We run two experiments where first we sweep the hidden layer size from 1 to 100 and second from 1 to 1000. Picking a small value such as 100 allows us to see if the linear search would be faster than the binary search due to overhead that comes from the modifications that we do in the binary search. Such small number also allows us to see if there is a global maximum in the curve fit that we do when we plot the accuracies over the hidden layer units. When we run the same experiment with 1000, it allows us to see real improvement that binary search provides. When we run sweep from 1 to 100, we fail to observe any global maximum that would generate a curve with a single global maximum in our Figures \ref{fig:titanic_accuracy_100}, \ref{fig:titanic_accuracy_100_with_dropout}, \ref{fig:churn_accuracy_100} and \ref{fig:churn_accuracy_100_with_dropout}. Absence of such global maximum fails the assumptions that we explained for the binary search to be applied, therefore, we cannot get global optimal solution via binary search for the experiment where N is 100. This proves that the binary search would fail to find the global maximum when the assumptions are not met. However, when we run the experiment with N is 1000, we observe the existence of a global maximum (albeit with the requirement of excluding the first few points), and the binary search proves to find the global maximum 100 times faster than the linear search in Figures \ref{fig:titanic_accuracy_1000}, \ref{fig:titanic_accuracy_1000_with_dropout}, \ref{fig:churn_accuracy_1000} and \ref{fig:churn_accuracy_1000_with_dropout}. In \autoref{table:modelsearching}, for runs where iteration count is 100, we report the best accuracies and architectures found by linear search since binary search fails and can only find suboptimal accuracies. For runs where iteration count is 1000, linear and binary search find the same architectures so the reported best model accuracies are results of both search methods. Overall, the experiments agree with the assumptions that we laid out in the Section \ref{binarysearch}, i.e binary search risks missing the global optimum if the conditions are not met, however when the assumptions are correct, it finds the global maximum several orders of magnitude faster than the naive approach. 
Additionally, when we apply a dropout layer to the models, for the case of Titanic 100 (Figure \ref{fig:titanic_accuracy_100_with_dropout}) and Churn 1000 (Figure \ref{fig:churn_accuracy_1000_with_dropout}), we see a decrease in the accuracies for about 3\%. Also, the best architecture candidates found are about 35\% smaller and the binary search converges faster when the dropout layer was included to the Titanic and Churn models (Figures \ref{fig:titanic_accuracy_1000_with_dropout} and \ref{fig:churn_accuracy_1000_with_dropout}).   

\section{Conclusion}\label{conclusion}

In this paper, we propose a framework to find the optimal architecture size for binary classification problems. We employ linear search and binary search to find such architecture size to give the highest accuracy. We use the Titanic dataset and the Churn Rate dataset and report a 100x improvement in finding the best model architecture when we apply the modified binary search compared to the linear search. We also show what happens when the assumptions that we lay out for binary search are not met. Binary search fails to find the global maximum solution and is stuck on a local solution.

\section{Future Work}\label{application}

In this study, we focused on datasets that can be modeled as binary classification problems where the output layer is 0 or 1. In the future, investigating these methods on multi-class classification problems or on models with more than one hidden layer can be worthwhile. 

Additionally, in this paper, we assumed that the accuracy graph when plotted against the architecture size would have a convex shape leading to a global maximum. In the future, we can look into removing this assumption in order to generalize the approach to all datasets.

\section*{Acknowledgment}

We would like to acknowledge Drexel Society of Artificial Intelligence for its contributions and support for this research.

\end{document}